\documentclass[a4paper,man,natbib]{apa6}
\usepackage{setspace}
\usepackage{array}
\usepackage[english]{babel}
\usepackage[utf8x]{inputenc}
\usepackage[colorinlistoftodos]{todonotes}
\usepackage{amsmath,amssymb,amsfonts,amsthm,bm}
\usepackage{graphicx}
\usepackage{pslatex}
\usepackage{caption}
\usepackage{subcaption}
\usepackage{siunitx}

\title{\textbf{Evaluating (and improving) the correspondence} \protect\\ \textbf{between deep neural networks and human representations}}
\shorttitle{Deep networks and human representations}

\fourauthors{Joshua C. Peterson}{Joshua T. Abbott}{Thomas L. Griffiths}{\par \bigskip *Corresponding author}
\fouraffiliations{Department of Psychology, University of California, Berkeley}{Department of Psychology, University of California, Berkeley}{Department of Psychology, University of California, Berkeley}
{Email: peterson.c.joshua@gmail.com\\
\singlespacing Department of Psychology\\
2121 Berkeley Way\\
University of California, Berkeley\\
Berkeley, CA 94720-1650}
\authornote{\noindent This work was supported by grant number FA9550-13-1-0170 from the Air Force Office of Scientific Research and grant number 1718550 from the National Science Foundation. Preliminary results (the animals domain from Experiment 1) were presented at the 38th Annual Conference of the Cognitive Science Society \citep{JoshJosh&TomDeepSim2016}.}

\abstract{Decades of psychological research have been aimed at modeling how people learn features and categories. The empirical validation of these theories is often based on artificial stimuli with simple representations. Recently, deep neural networks have reached or surpassed human accuracy on tasks such as identifying objects in natural images. These networks learn representations of real-world stimuli that can potentially be leveraged to capture psychological representations. We find that state-of-the-art object classification networks provide surprisingly accurate predictions of human similarity judgments for natural images, but fail to capture some of the structure represented by people. We show that a simple transformation that corrects these discrepancies can be obtained through convex optimization. We use the resulting representations to predict the difficulty of learning novel categories of natural images. Our results extend the scope of psychological experiments and computational modeling by enabling tractable use of large natural stimulus sets.
}

\begin{document}
\maketitle

\setcounter{secnumdepth}{3}

\section{Introduction}
Humans possess a remarkable ability to cope with complex inductive problems in the natural world. For this reason, trying to understand how people solve these problems has been one of the core programs of cognitive science for decades.  Despite considerable theoretical progress, experimental validation has been limited largely to laboratory settings with artificial stimuli with simple representations (e.g., strings of binary digits, colored shapes; although for a recent exception see \citealp*{meagher2017organized}). Natural stimuli such as large sets of realistic images of animals will require a complex representation that may be difficult to easily interpret or manipulate in the lab. Psychologists have provided clever workarounds to this problem by inferring representations of a set of stimuli directly from human generalization data \citep{shepard1980multidimensional}, but only a relatively small set of stimuli can be compared in an experiment and novel stimuli cannot be incorporated. This makes it nearly impossible to identify representations for all of the myriad stimulus variability in the natural world, or even a small chunk of it.

Deep neural networks (DNNs) have been shown to approach or exceed human performance in a number of key perceptual tasks such as object categorization and scene understanding \citep{lecun2015deep}, among other breakthroughs in natural language processing \citep{collobert2011natural} and reinforcement learning \citep{mnih2015human}. These networks can be trained on millions of images, allowing them to learn sets of features that generalize broadly and solve real problems. In this paper, we explore how well the representations discovered by DNNs align with human psychological representations of natural images, show how they can be adjusted to increase this correspondence, and demonstrate that the resulting representations can be used to predict complex human behaviors such as learning novel categories.

Following the success of DNNs in computer vision, recent work has begun to compare the properties of these networks to psychological and neural data. 
Much of the initial work in comparing deep neural network representations to those of humans comes from neuroscience. For example, early work found that neural network representations beat out $36$ other popular models from neuroscience and computer vision in predicting IT cortex representations \citep{khaligh2014deep}, and later work found a similar primacy of these representations in predicting voxel-wise activity across the visual hierarchy \citep{agrawal_pixels_2014}. However, neural representations are not necessarily the gold standard for capturing all of the complex structure of human mental representations. Human similarity judgments for a set of objects encode representational detail that cannot be estimated by inferotemporal cortex representations, which are more similar to monkey inferotemporal cortex than to human psychological representations \citep{mur_human_2013}. For this reason, estimating human behavior directly may also be fruitful, and possibly more informative. Several recent studies have seen some initial success in applying representations from deep neural networks to psychological tasks, including predicting human typicality ratings \citep{lake_deep_2015} and memorability \citep{ICCV15_ObjectMemorability} for natural object images. More recently, it was shown that human shape sensitivity for natural images could be explained well for the first time using deep neural networks \citep{kubilius2016deep}, which now constitute a near essential baseline for emerging models of human shape perception \citep{erdogan2017visual}. A follow-up to our own previous work \citep{JoshJosh&TomDeepSim2016} showed that important categorical information is missing from deep representations \citep{jozwik2017deep}.

Comparing the representations formed by deep neural networks with those used by people is challenging, as human psychological representations cannot be observed directly. Our approach is to solve this problem by exploiting the close relationship between \textit{representation} and \textit{similarity} (i.e., every similarity function over a set of pairs of data points corresponds to an implicit representation of those points). This provides an empirical basis for the first detailed evaluation of DNNs as an approximation of human psychological representations. We subject both DNN and human similarities to an ensemble of classic psychological methods for probing the spatial and taxonomic information they encode. This identifies aspects of human psychological representations that are captured by DNNs, but also significant ways in which they seem to differ. We then consider whether a better model of human representations can be efficiently bootstrapped by transforming  the deep representations. The resulting method opens the door to ecological validation of decades of psychological theory using large datasets of highly complex, natural stimuli, which we demonstrate by predicting the difficulty with which people learn natural image categories.

\section{Experiment 1: Evaluating the correspondence between representations}

Human psychological representations are not directly observable, and cannot yet be inferred from neural activity \citep{mur_human_2013}. However, psychologists have developed methods for inferring representations from behavior alone. Human similarity judgments capture stimulus generalization behavior \citep{shepard_toward_1987} and have been shown to encode the complex spatial, hierarchical \citep{shepard1980multidimensional}, and overlapping \citep{shepard1979additive} structure of psychological representations, around which numerous psychological models of categorization and inference are built \citep{goldstone1994role,nosofsky1987attention,kruschke1992alcove}. If we can capture similarity judgments, we will have obtained a considerably high resolution picture of human psychological representations. In Experiment 1 we evaluated the performance of deep neural networks in predicting human similarity judgments for six large sets of natural images drawn from a variety of visual domains: animals, automobiles, fruits, furniture, vegetables, and a set intended to cross-cut visual categories (which we refer to below as ``various''). 

\subsection{Methods}

\subsubsection{Stimuli}
\begin{figure}[!h]
\begin{center}
\includegraphics[width=0.7\linewidth,keepaspectratio]{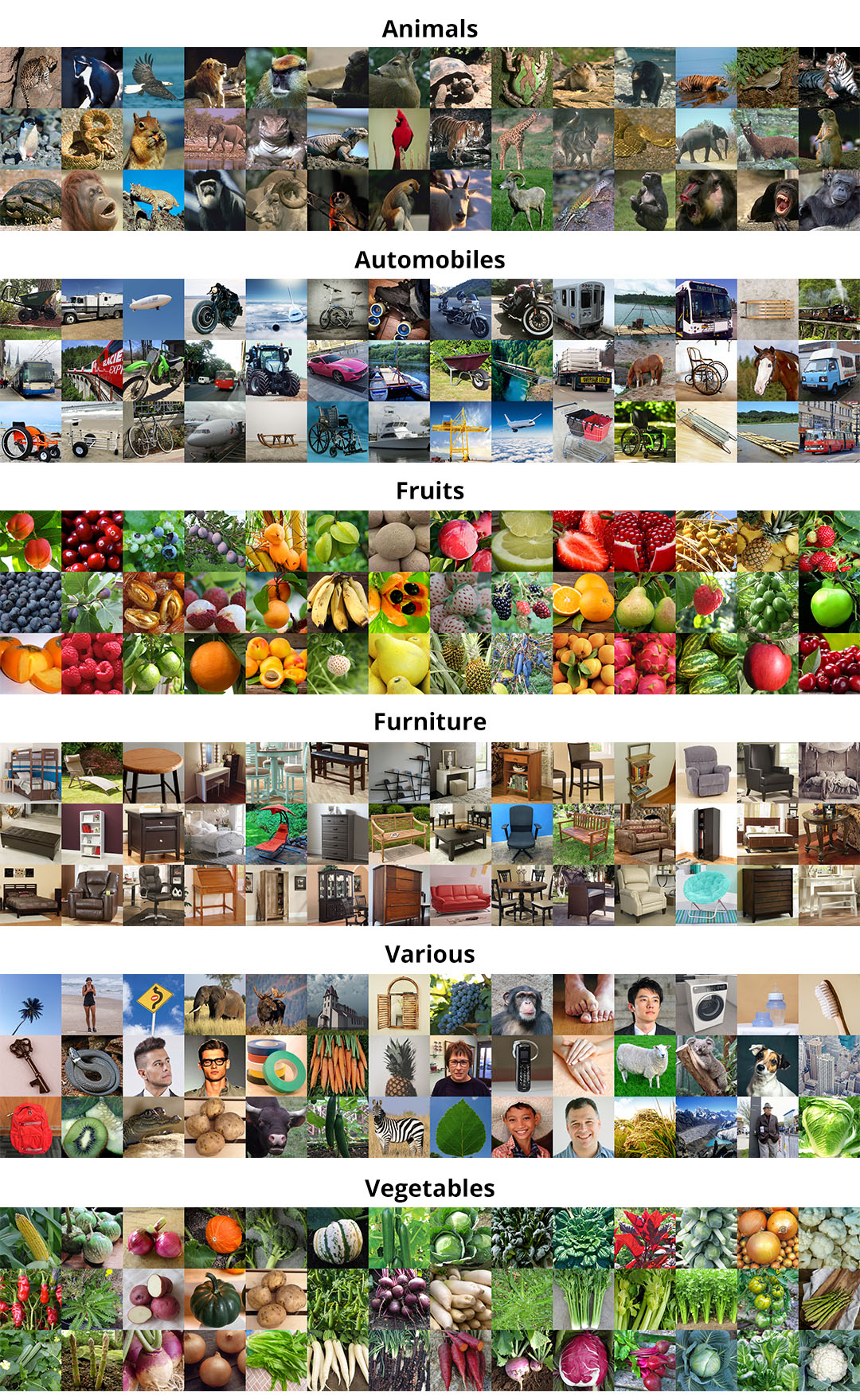}
\end{center}
\caption{Example image stimuli from our six domains.}
\label{fig:image-samples}
\end{figure}

Stimuli were hand-collected for each of the six domains, digital photos that were meant to exhibit wide variety in object pose, camera viewpoint, formality, and subordinate class. Each domain contained 120 total images, each cropped and resized to $300\times300$ pixel dimensions. An example subset of these images for each dataset is provided in Fig. \ref{fig:image-samples}.

\subsubsection{Procedure}

For all six stimulus categories, we collected pairwise image similarity ratings (within each category) from human participants on Amazon Mechanical Turk. Participants were paid $\$0.02$ to rate the similarity of four pairs of images within one of the six categories on a scale from $0$ (``not similar at all'') to $10$ (``very similar''). They could repeat the task as many times as they wanted, but we did not allow repeat ratings of the same unique image pair. We obtained exactly 10 unique ratings for each pair of images ($7,140$ total) in each category, yielding $71,400$ ratings per category ($428,400$ total ratings), from over $1,200$ unique participants. The result is six $120\times120$ similarity matrices after averaging over individual judgments, for which each entry represents human psychological similarity between a pair of objects.

\subsubsection{Deep neural network representations}

When deep neural networks are presented with an image, the nodes that comprise the network obtain different activation values. We can take these activation values as a vector of ``features'' representing the image. These feature vectors can be collected into a feature matrix ${\bf F}$, which specifies a multidimensional feature representation (columns) for each image (rows). A similarity matrix ${\bf S}$, in which the entry $s_{ij}$ gives the similarity between images $i$ and $j$, can then be approximated by the matrix product
\begin{equation}
    {\bf S} = {\bf F} {\bf F}^T , \label{eq:inner}
\end{equation}
modeling $s_{ij}$ as the inner product of the vectors representing images $i$ and $j$. Given human similarity judgments ${\bf S}$ and an artificial feature representation ${\bf F}$, we can evaluate the correspondence between the two by computing the correlation between the entries in ${\bf S}$ and ${\bf F}{\bf F}^T$. 

For each image in all six categories, we extracted deep feature representations using four highly popular convolutional neural network image classifiers that were pretrained in Caffe \citep{jia2014caffe} on \textsc{ILSVRC12}, a large dataset of $1.2$ million images taken from 1000 objects categories in the ImageNet database \citep{deng2009imagenet}. This dataset serves as a central benchmark in the computer vision community. Our own image datasets were not explicitly sampled from categories in \textsc{ILSVRC12} and likely diverge to some degree. For example, of the 1000 \textsc{ILSVRC12} classes, 120 are different dog breeds, whereas our animal set contains no dogs. The networks, in order of depth, are AlexNet \citep{krizhevsky2012imagenet}, VGG \citep{simonyan2014very}, GoogLeNet \citep{szegedy2014going}, and ResNet \citep{he2016deep}, three of which are \textsc{ILSVRC12} competition winners. VGG, GoogLeNet, and ResNet all achieve at least half the error rate of AlexNet. Images are fed forward through each network as non-flattened tensors, and activations are recorded at each layer of the network. For most of our analyses besides the AlexNet layer analysis, we extract only the activations at the final hidden layer of each network. For AlexNet and VGG, this is a $4096$-dimensional fully-connected layer, while the last layers in GoogleNet and ResNet are 1024- and 2048-dimensional pooling layers respectively. As an example, feature extraction for the animals training image set provides a $120\times4096$ matrix. All feature sets were then z-score normalized. Beyond these classification networks, we also included a very recent state-of-the-art unsupervised deep image network \citep{donahue2016adversarial,dumoulin2016adversarially}, a generative model trained to model the distribution of the entire \textsc{ILSVRC12} dataset. This network (BiGAN) is a bidirectional variant of a Generative Adversarial Network \citep{goodfellow2014generative} that can both generate images from a uniform latent variable and perform inference to project real images into this latent space. We use the 200-dimensional projections into this latent space as our representation for this network. As an additional baseline, we also extract two forms of shallow (non-deep) features using previously popular methods from computer vision called the Scale-invariant feature transform (SIFT) \citep{lowe2004distinctive}, using the bag-of-words technique trained on a large image database, and Histogram of Oriented Gradients (HOG) \citep{dalal2005histograms}, with a bin size of $2\times2$.

\subsection{Results and Discussion}

\begin{figure}[!t]
\begin{center}
\includegraphics[width=0.8\linewidth,keepaspectratio]{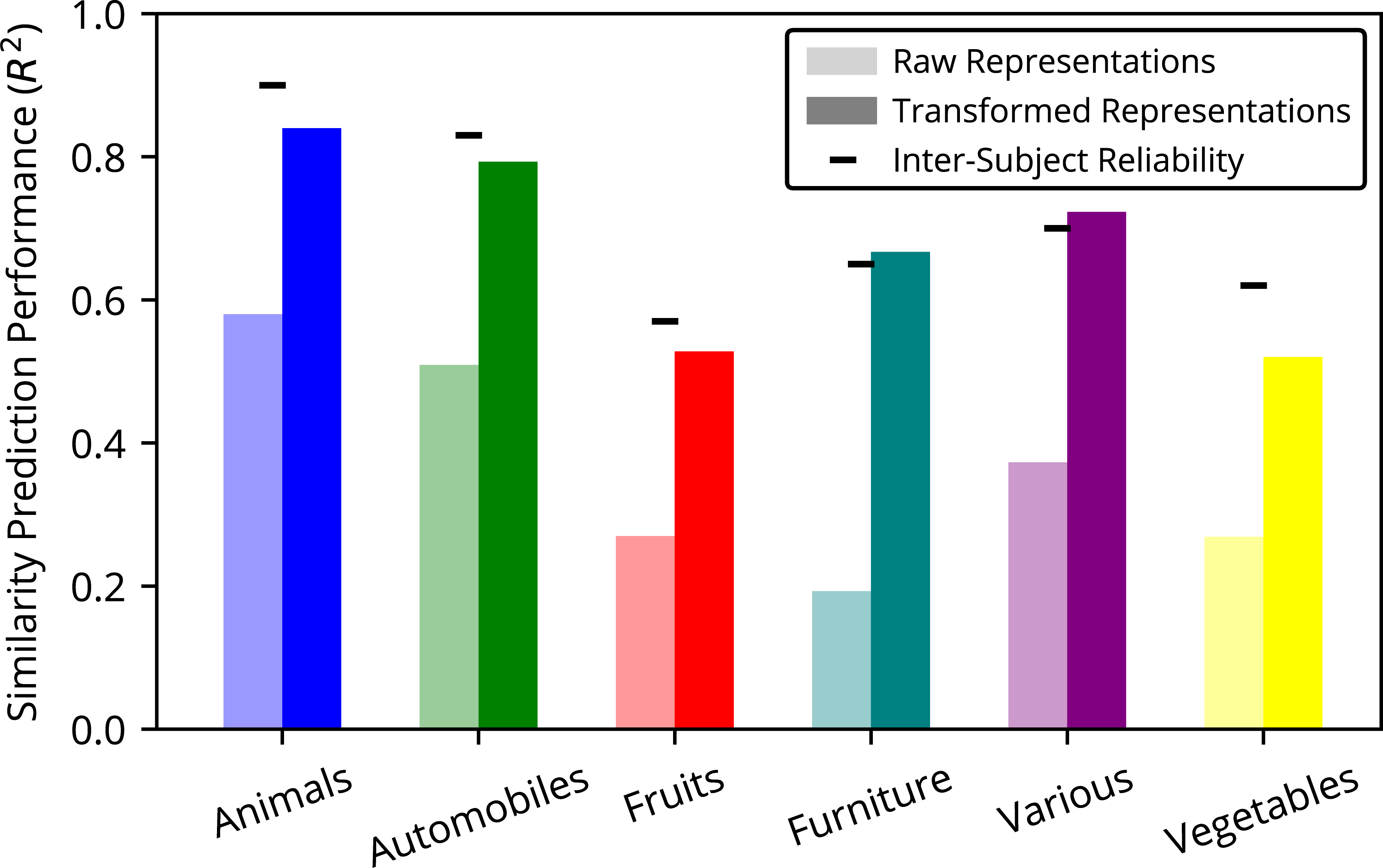}
\end{center}
\caption{Model performance (proportion of variance accounted for, $R^2$) in predicting human similarity judgments for each image set using the best raw (light colors) and best transformed (dark colors) DNN representations.}
\label{fig:raw-vs-transformed}
\end{figure}

We began our analyses by computing the correlation between the human similarity judgments and the inner products computed in the deep feature representations.
The variance explained in human similarity judgments by the best performing DNN architecture (this was VGG in all cases) is plotted in Fig. \ref{fig:raw-vs-transformed} (lighter colors) and given in Table \ref{results-and-controls} (``raw''), and indicates that the raw deep representations can give reasonable first approximation to human similarity judgments. We found that alternative metrics such as Euclidean distance yielded essentially identical results (not shown).

\begin{table}[]
\centering
\caption{Variance explained in human similarity judgments for raw and transformed representations for the best performing network (VGG).}
\label{results-and-controls}
\begin{tabular}{lcccc}
Dataset     & Raw $R^2$ & Transformed $R^2$ & CV Control $R^2$ & Human Inter-reliability \\
\midrule
Animals     & 0.58 & 0.84 & 0.74 & 0.90 \\
Automobiles & 0.51 & 0.79 & 0.58 & 0.83 \\
Fruits      & 0.27 & 0.53 & 0.36 & 0.57 \\
Furniture   & 0.19 & 0.67 & 0.35 & 0.65 \\
Various     & 0.37 & 0.72 & 0.54 & 0.70 \\
Vegetables  & 0.27 & 0.52 & 0.35 & 0.62 \\
\bottomrule
\end{tabular}
\end{table}

To better understand how DNNs succeed and fail to reproduce the structure of psychological representations, we applied two classic psychological tools: non-metric multidimensional scaling, which converts similarities into a spatial representation, and hierarchical clustering, which produces a tree structure (dendrogram) \citep{shepard1980multidimensional}. 
For our NMDS analysis, we used the scikit-learn Python library to obtain only two-dimensional solutions, with a maximum iteration limit of $10,000$ in fitting the models through gradient descent, and a convergence tolerance of 1e-100. Embeddings were first initialized with standard metric MDS, and we took the best fitting solution of four independent initializations. For HCA, we used the scipy Python library, with a centroid linkage function in all models.

\begin{figure}[!t]
\begin{center}
\includegraphics[width=1.0\linewidth,keepaspectratio]{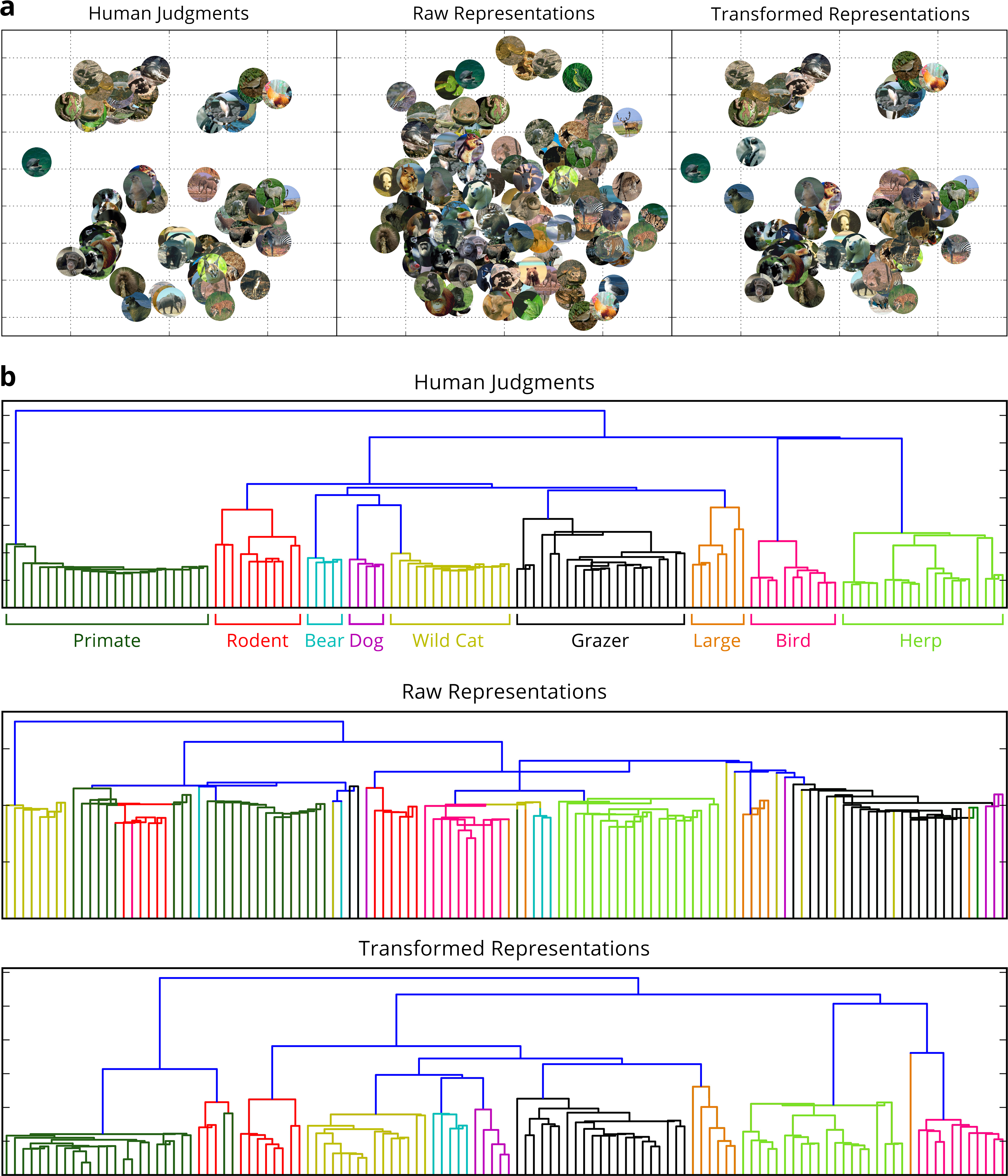}
\end{center}
\caption{Representations of Animals. (a) Non-metric multidimensional scaling solutions for human similarity judgments (left), raw DNN representations (middle), and transformed DNN representations (right). (b), Dendrograms of hierarchical clusterings (centroid method) for human similarity judgments (top), raw DNN representations (middle), and the transformed DNN representations (bottom).} 
\label{fig:animal-reps}
\end{figure}

The results for the best-performing DNN on the animals stimuli are shown in Fig. \ref{fig:animal-reps}, and point out the most crucial differences in these two representations. Human representations exhibit highly distinguished clusters in the spatial projections and intuitive taxonomic structure in the dendrograms, neither of which are  present in the DNN representations. This gives us an idea of what relevant information is missing from the deep representations in order to fully approximate human representations.

\begin{figure}[!h]
\begin{center}
\includegraphics[width=0.8\linewidth,keepaspectratio]{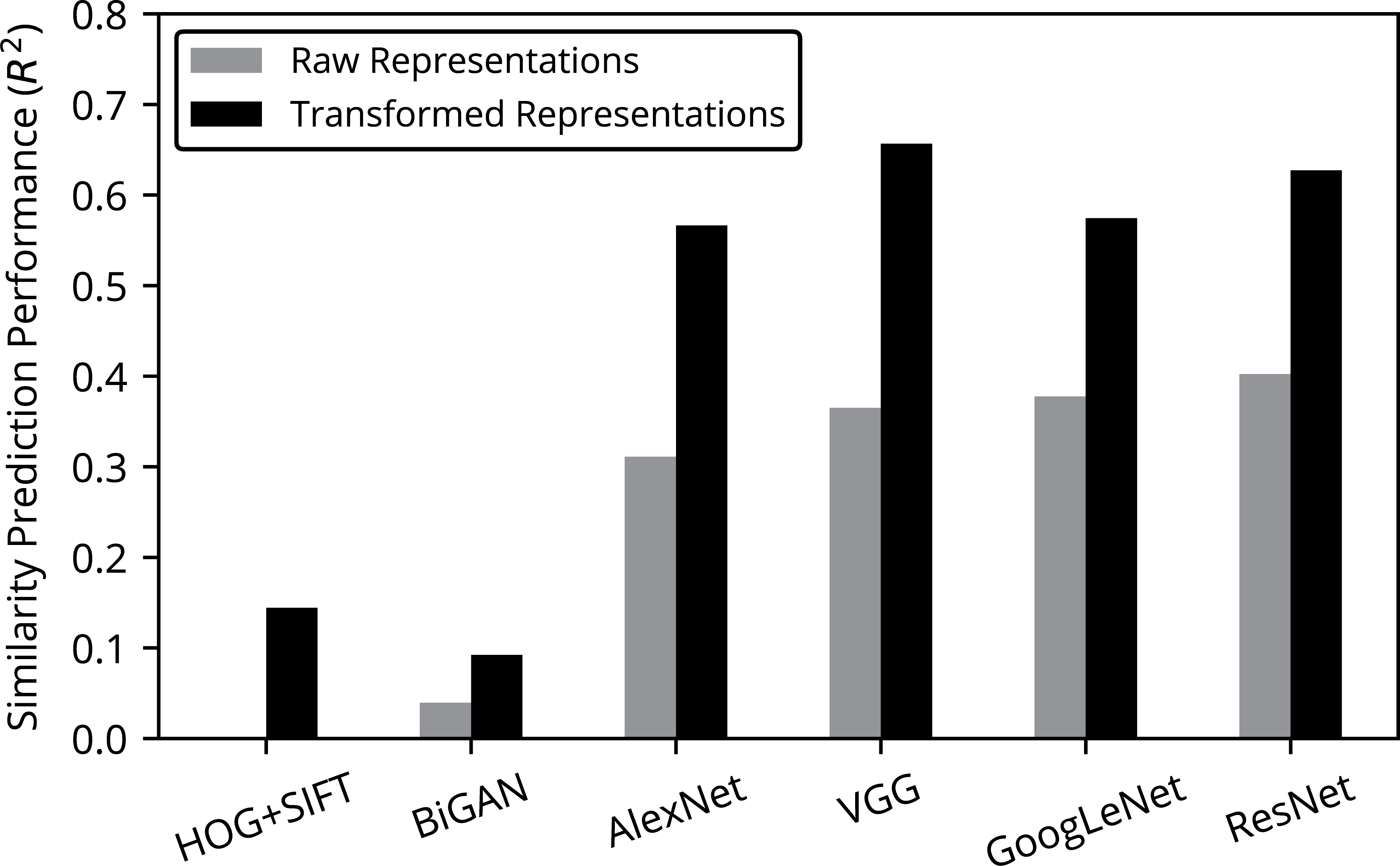}
\end{center}
\caption{Similarity prediction performance using the best weighted representations from four popular deep classifiers, an unsupervised network (BiGAN), and a non-deep baseline (HOG+SIFT). Results are averaged across all six image sets.}
\label{fig:network-comparison}
\end{figure}

Beyond identifying the DNN that best captures human similarity judgments, we wanted to understand how competing networks compare in their predictive ability. Fig. \ref{fig:network-comparison} shows the results of comparing the representations from all four classification networks, as well as a recent high-performing unsupervised deep architecture \citep[BiGAN;][]{donahue2016adversarial,dumoulin2016adversarially} and two older, non-deep standards from computer vision: HOG \citep{lowe2004distinctive} and SIFT \citep{dalal2005histograms} features. We find that most classification networks perform similarly, yet VGG is slightly better on average. Surprisingly, representations from the BiGAN, while useful for machine object classification \citep{donahue2016adversarial}, don't seem to correspond as well to human representations, and are even less effective than shallow methods like HOG+SIFT.

\begin{figure}[!h]
\begin{center}
\includegraphics[width=0.8\linewidth,keepaspectratio]{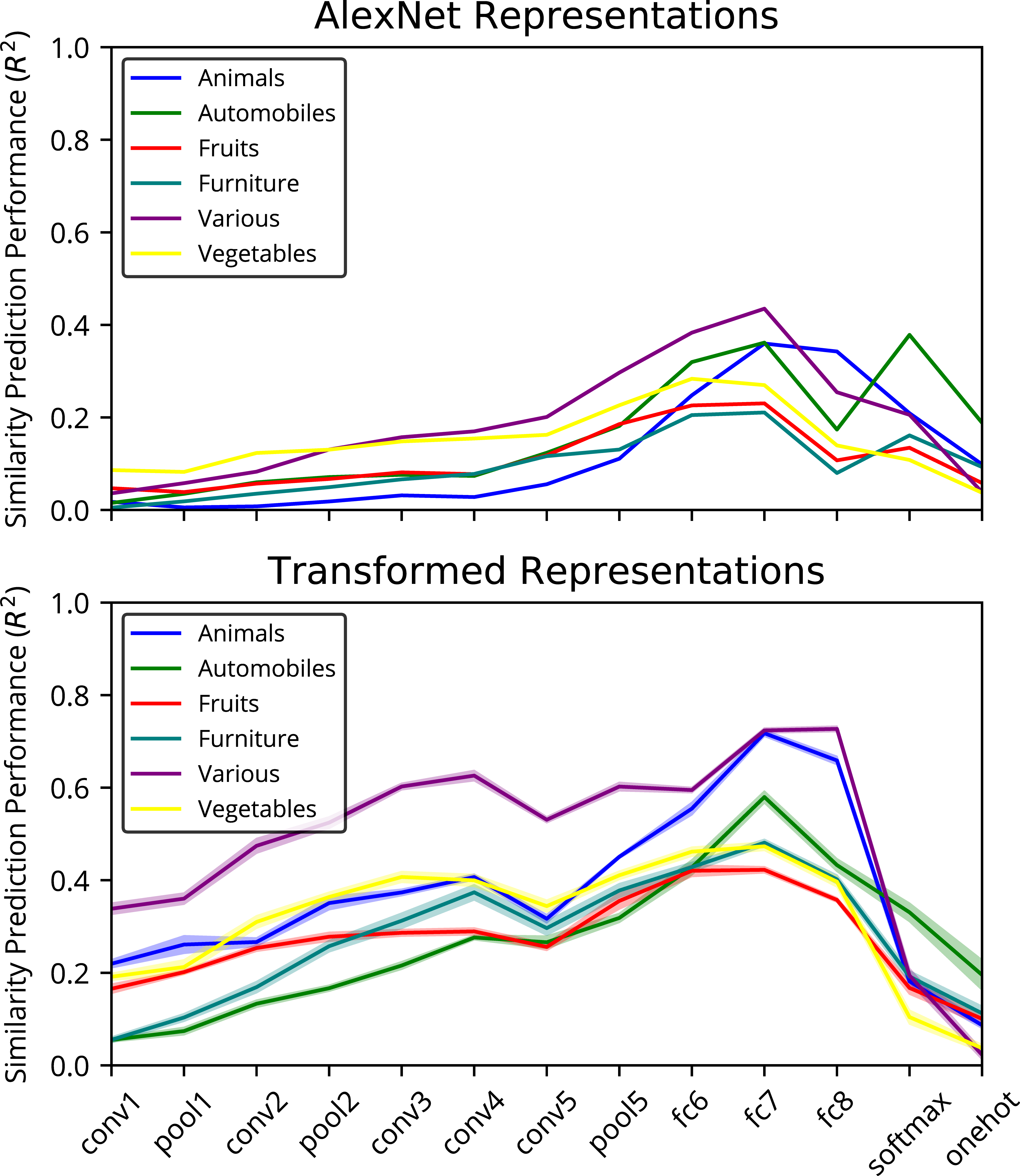}
\end{center}
\caption{Similarity prediction performance using transformed representations at each layer of AlexNet for each dataset (``softmax'' is predicted class probabilities, and ``one-hot'' is predicted class labels).}
\label{fig:layer-comparison}
\end{figure}

Additionally, using AlexNet, which has a manageable yet still large number of layers, we examined performance at each layer of the network, including final class probabilities and discrete labels. As Fig. \ref{fig:layer-comparison} shows, performance climbs as the depth of the network increases, but falls off near the end when the final classification outputs near. For all datasets, the best layer was the final hidden layer, yielding a $4096$-dimensional vector, as opposed to the classification layer which by design must shrink to merely $1000$ dimensions. This indicates that relatively high-level, yet non-semantic information is most relevant to the human judgments we obtained.

\section{Transforming deep representations}

Experiment 1 showed that the raw representations discovered by deep neural networks perform reasonably well as predictors of human similarity judgments. This correspondence suggests that deep neural networks could potentially provide an indispensable tool to  psychologists aiming to test theories with naturalistic stimuli. Even a crude approximation of a complex representation may vastly outperform classic low-level features often used to characterize natural stimuli (e.g., Gabor wavelet responses). More importantly, having a representation that approximates human similarity judgments provides a starting point for identifying representations that are even more closely aligned with people's intuitions. In this section, we explore how DNN representations can be transformed to increase the alignment with psychological representations. 

\subsection{Transforming representations}

The model of similarity judgments given in Equation \ref{eq:inner} can be augmented with a set of weights on the features used to compute similarity, with 
\begin{equation}
{\bf S} = {\bf F} {\bf W} {\bf F}^T , \label{eq:weighted}
\end{equation}
where ${\bf W}$ is a diagonal matrix of dimension weights. This formulation is similar to that employed by additive clustering models \citep{shepard1979additive}, wherein $\mathbf{F}$ represents a binary feature identity matrix, and is similar to Tversky's classic model of similarity \citep{tversky1977features,navarro2004common}. Concretely, it provides a way to specify the relationship between a feature representation and stimulus similarities. When used with continuous features, this approach is akin to factor analysis.

Given an existing feature-by-object matrix $\mathbf{F}$, we can show that the diagonal of $\mathbf{W}$, the vector of weights ${\bf w}$, can be expressed as the solution to a linear regression problem where the predictors for each similarity $s_{ij}$ are the (elementwise) product of the values of each feature for objects $i$ and $j$ (i.e. each row of the regression design matrix ${\bf X}$ can be written as $\mathbf{F_i}\circ{\mathbf{F_j}}$, where $\circ$ is the Hadamard product). The similarity $s_{ij}$ between objects $i$ and $j$ is therefore modeled as $s_{ij} = \sum_k w_k f_{ik}f_{jk}$, where $f_{ik}$ is the $k$th feature of image $i$ and $w_k$ is its weight. The squared error in reconstructing the human similarity judgments can be minimized by convex optimization. \cite{gershman2015phrase} proposed a similar method using a full $\mathbf{W}$ matrix, which is a more expressive model, but requires fitting more parameters. We use a diagonal $\mathbf{W}$ matrix to minimize the amount of data and regularization needed to fit our models, and assume that the needed transformation is as simple as possible.

The resulting alignment method is akin to metric learning methods in machine learning \citep{kulis2013metric}. Estimating both the features and the weights that contribute to human similarity judgments, even for simple stimuli, is a historically challenging problem \citep{shepard1979additive}. Our main contribution is to propose that $\mathbf{F}$ be substituted by features from a deep neural network, and only ${\bf w}$ be learned. This both coheres with our comparison framework and greatly simplifies the problem of estimating human representations. 

If ${\bf w}$ is also constrained to be nonnegative, then the square root of these weights can be interpreted as a multiplicative rescaling of the features. This makes it possible to directly construct transformed spatial representations of stimuli. Since a direct feature transformation is not necessary for our evaluation, we include no such constraint in the results that follow. However, it should be noted that this variation allows for applications where it is essential that transformed features be exposed (i.e., when similarities will not suffice).

\subsection{Learning the transformations}

Freely identifying the ${\bf w}$ that best predicts human similarity judgments runs the risk of overfitting, since our DNNs generate thousands of features. To address this, all of our models use L2 regularization on ${\bf w}$, penalizing models for which the inner product ${\bf w}^T{\bf w}$ is large. If we minimize the squared error in the reconstruction of $s_{ij}$ with L2 regularization on ${\bf w}$, the result is a convex optimization problem that is equivalent to ridge regression \citep{friedman2001elements}. Given the size of the problem, we find ${\bf w}$ by gradient descent on an objective function combining the squared error and ${\bf w}^T{\bf w}$, with the latter weighted by a regularization parameter $\lambda$. To accomplish this, we used the ridge regression implementation in the scikit-learn Python library with a stochastic average gradient solver. We use $6$-fold cross-validation to find the best value for this regularization parameter, optimizing generalization performance on held-out data. We chose $6$ folds as a rule of thumb, although the results did not appear to be largely dependent on the number of folds used. We report variance explained only for models predicting non-redundant similarity values (only the lower triangle of the similarity matrix, excluding the diagonal).

\subsection{Improvements through feature adaptation}

We applied the method for adapting the DNN representations outlined above to the human similarity judgments and network representations used in Experiment 1. The best $\lambda$ values for each dataset were comparable, in the range of $2000-9000$. After learning the best cross-validated weights ${\bf w}$ that map these features to human similarity judgments, the new representation that emerges explained nearly twice the variance for all datasets after cross-validating predictions (Figs. \ref{fig:raw-vs-transformed} and \ref{fig:network-comparison}, darker colors). We also provide the raw scores for the best performing model (VGG) in Table \ref{results-and-controls}, along with the results of a control cross-validation (``CV Control'') scheme in which no single images occurred in both the training fold sets and test folds (as opposed to exclusivity with respect only to \textit{pairs} of images). The MDS and dendrogram plots for the transformed representations in Fig. \ref{fig:animal-reps} show a strong resemblance to the original human judgments. Notably, taxonomic structure and spatial clustering is almost entirely reconstructed, effectively bridging the gap between human and deep representations.

\subsection{Additional baseline models}
As additional check for overfitting, we constructed baseline models for each set of deep representations for each image dataset in which either (1) the rows, (2) the columns (separately for each row), or (3) both row and columns of the regression design matrix ${\bf X}$ were randomly permuted. The order of the target similarities from $S$ remained unchanged. When all three models were subject to the same cross-validation procedure as the unshuffled models, variance explained ($R^2$) never reached or exceeded $0.01$. This confirms that our regularization procedure was successful in controlling overfitting.

\subsection{Inter-domain transfer}
The transformations learned are highly contingent on the domain, and do not generalize well to others (e.g., a transformation trained on fruits is not effective when tested on animals). Table \ref{transfer-results} shows the performance of the best DNN representations for each domain when applied to each other domain. The correlations are relatively poor, and worse than those produced by the best untransformed representations.

\begin{table}[]
\centering
\caption{Inter-domain generalization of best performing DNN transformations}
\label{transfer-results}
\begin{tabular}{lll}
Training Set & Test Set        & $R^2$ \\
\midrule
Animals     & Fruits      & 0.11  \\
Animals     & Furniture   & 0.02  \\
Animals     & Vegetables  & 0.11  \\
Animals     & Automobiles & 0.17  \\
Animals     & Various     & 0.12  \\
Fruits      & Animals     & 0.14  \\
Fruits      & Furniture   & 0.12  \\
Fruits      & Vegetables  & 0.14  \\
Fruits      & Automobiles & 0.25  \\
Fruits      & Various     & 0.13  \\
Furniture   & Animals     & 0.20  \\
Furniture   & Fruits      & 0.07  \\
Furniture   & Vegetables  & 0.11  \\
Furniture   & Automobiles & 0.10  \\
Furniture   & Various     & 0.06  \\
Vegetables  & Animals     & 0.30  \\
Vegetables  & Fruits      & 0.10  \\
Vegetables  & Furniture   & 0.11  \\
Vegetables  & Automobiles & 0.21  \\
Vegetables  & Various     & 0.08  \\
Automobiles & Animals     & 0.36  \\
Automobiles & Fruits      & 0.11  \\
Automobiles & Furniture   & 0.07  \\
Automobiles & Vegetables  & 0.13  \\
Automobiles & Various     & 0.12  \\
Various     & Animals     & 0.41  \\
Various     & Fruits      & 0.05  \\
Various     & Furniture   & 0.06  \\
Various     & Vegetables  & 0.11  \\
Various     & Automobiles & 0.21  \\ 
\bottomrule
\end{tabular}

Note: Comparison $R^2$ values for best performing networks in each domain appear in Table \ref{results-and-controls}.
\end{table}

This pattern of poor inter-domain transfer is to be expected, since the number of DNN features is large and each domain only covers a small subset of the space of images and thus only provides information about the value of a small subset of features. However, it is possible to use the same adaptation method to produce a more robust transformation of the DNN representations for the purposes of predicting human similarity judgments. To do so, we learned a transformation using all six domains at once. This can also be thought of as a test of the robustness of our method when provided with an incomplete similarity matrix, specifically one containing only within-domain comparisons, yet still using all domains to constrain the ultimate model solution. This also allows for larger sets of images to be leveraged simultaneously for better learning.

We found this method to be highly effective, doubling the variance explained in human similarity judgments by the DNN representations from 30\% to 60\% after the transformation. A leave-one-out procedure in which every combination of five domains predicted the sixth provided similar improvements, as shown in Table \ref{generalization-results}. This is a strong control given that no images (and no similar images) are shared between the training and test sets in this formulation.

\begin{table}[]
\centering
\caption{Generalization performance leaving out a single domain and training on the remaining five.}
\label{generalization-results}
\begin{tabular}{lll}
Leave-out   & $R^2$ \\
\midrule
Animals     & 0.53  \\
Automobiles & 0.57  \\
Fruits      & 0.63  \\
Furniture   & 0.62  \\
Various     & 0.59  \\
Vegetables  & 0.63  \\
\bottomrule
\end{tabular}
\end{table}

\section{Experiment 2: Predicting the difficulty of learning categories of natural images}

A simple linear transformation was able to adapt DNN representations to predict human similarity judgments at a level that is close to the inter-rater reliability. The transformed representation also corrected for the qualitative differences between the raw DNN representation and psychological representations. These results indicate that the rich features formed by DNNs can be used to capture psychological representations of natural images, potentially making it possible to run a much wider range of psychological experiments with natural images as stimuli.

The value of these representations for broadening the scope of psychological research can only be assessed by establishing that they generalize to new stimuli, and are predictive of other aspects of human behavior. To further explore the generalizability and applicability of this approach, we applied the learned transformation to the DNN representations (from VGG) of six new sets of unseen images drawn from the same domains and assessed the ease with which people could learn categories constructed from the raw and transformed similarities. 

The categories we used were constructed via $k$-means clustering based on either the raw or transformed similarities, ensuring that each category consisted of a coherent group of images as assessed by the appropriate similarity measure. Consequently, we should expect the ease of learning those categories to reflect the extent to which people's sense of similarity has been captured. In addition, traditional image features such as HOG+SIFT should make category learning more difficult than using DNN features, given the mismatch between representations observed in our previous analyses. 

\subsection{Methods}

\subsubsection{Stimuli}

\begin{figure}[]
\begin{center}
\includegraphics[width=0.8\linewidth,keepaspectratio]{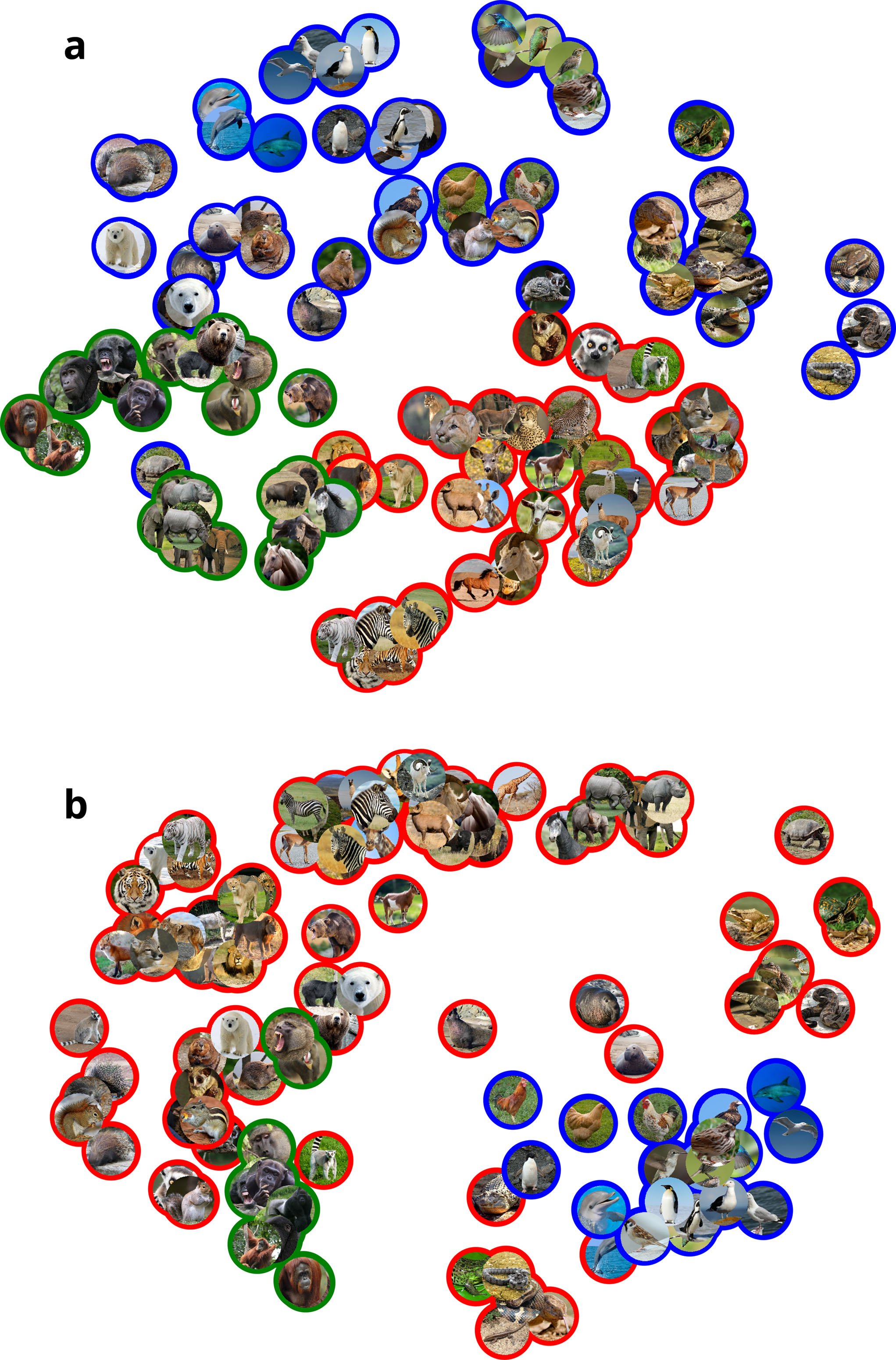}
\end{center}
\caption{Examples of animal clusterings used in our categorization experiments where $k=3$ for (a) the raw deep representations, and (b) the transformed deep representations. The transformation was learned on a different set of animal images, and appears to improve clustering in some aspects of the space. For example, the transformation makes primates more unique (i.e., not grouped with quadrupeds), and doesn't group small land and marine animals.} 
\label{fig:animal-clusters-example}
\end{figure}

Using the best performing network and layer for each image dataset, we applied the learned transformation to a second set of 120 new images in each category. This produced six predicted similarity matrices for each set. Using the rows of these matrices as image representations, we calculated $k$-means clusterings where the number of clusters ($k$) was either $2$, $3$, or $4$. We repeated this process using the untransformed representations, for which similarities were simply inner products. This resulted in the following between-subjects conditions for our experiment: space (transformed, raw) $\times$ $k$ (2,3,4) $\times$ domain (e.g., animals). We also replicated these experiments using baseline HOG+SIFT representations, yielding a total of 72 between-subjects conditions. An example of the clusterings used in the animal experiments where $k=3$ are shown in Fig. \ref{fig:animal-clusters-example}.

\subsubsection{Procedure}
A total of $2,880$ participants ($40$ per condition) were recruited on Amazon Mechanical Turk, paid \$$1.00$, and were not allowed to participate in multiple conditions. Participants in each condition were shown a single random sequence of the images from the dataset corresponding to their assigned condition and were instructed to press a key to indicate the correct category (where the correct category was the pre-defined cluster). Subjects could take as much time as they wanted to make their decisions. If a participant guessed incorrectly, an ``incorrect'' message was shown for $1.5$ seconds. If they guessed correctly, this message read ``correct''. Initially, participants performed poorly as they had little information to associate keys with clusters, but showed consistent progress after a few examples from each cluster. 

\subsection{Results and Discussion}

\begin{figure}[]
\begin{center}
\includegraphics[width=0.8\linewidth,keepaspectratio]{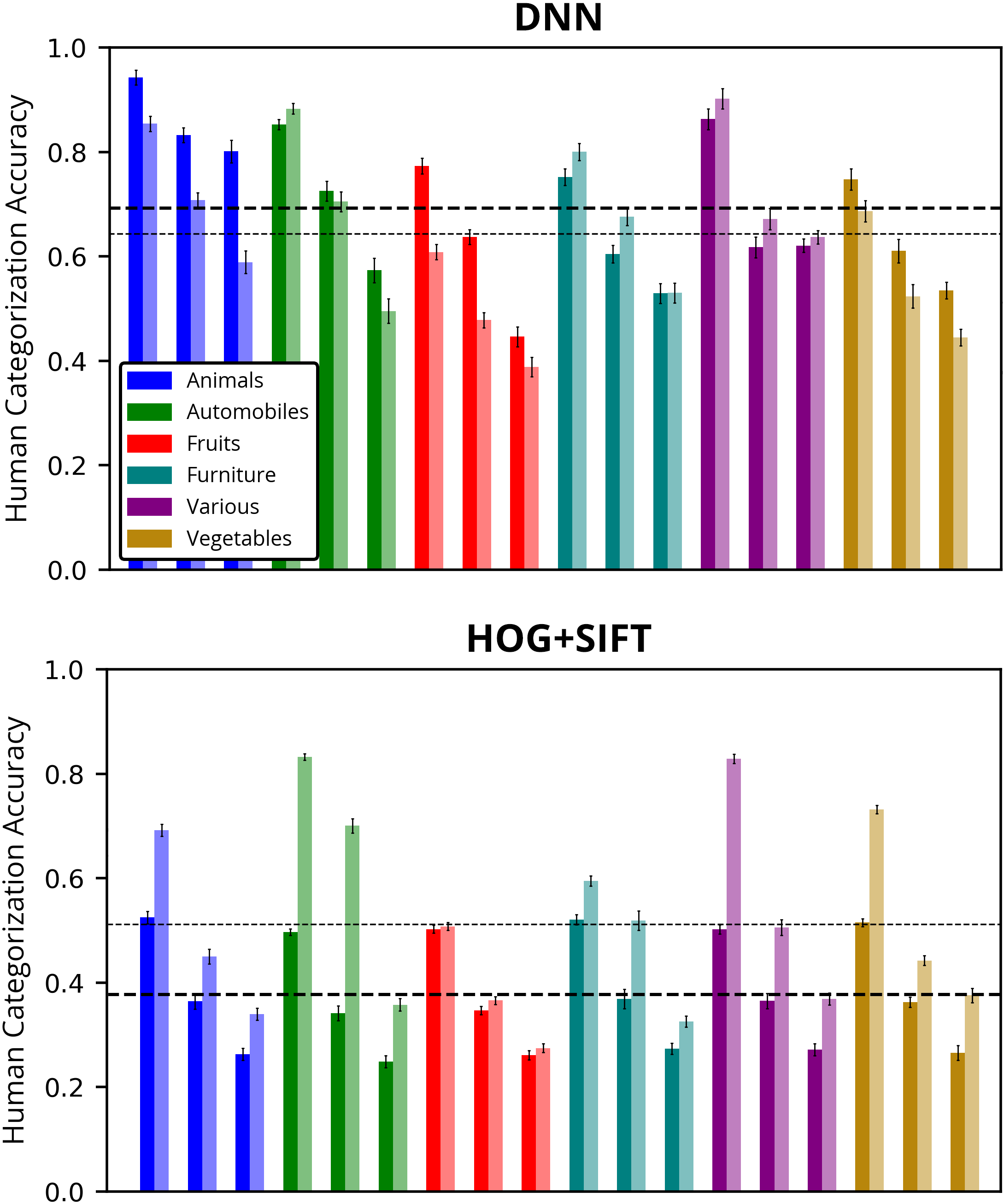}
\end{center}
\caption{Average human categorization performance on each image set using raw and transformed DNN representations (top) and baseline HOG+SIFT features (bottom). Darker colors represent transformed versions of the raw representations (lighter colors). The three sets of bars for each image set represent 2-, 3-, and 4-category versions of the experiment. Thick dashed lines represent average accuracy for the raw representations, and thick dashed lines represent average accuracy for the transformed representations.} 
\label{fig:cat-results-extended}
\end{figure}

\begin{figure}[]
\begin{center}
\includegraphics[width=0.8\linewidth,keepaspectratio]{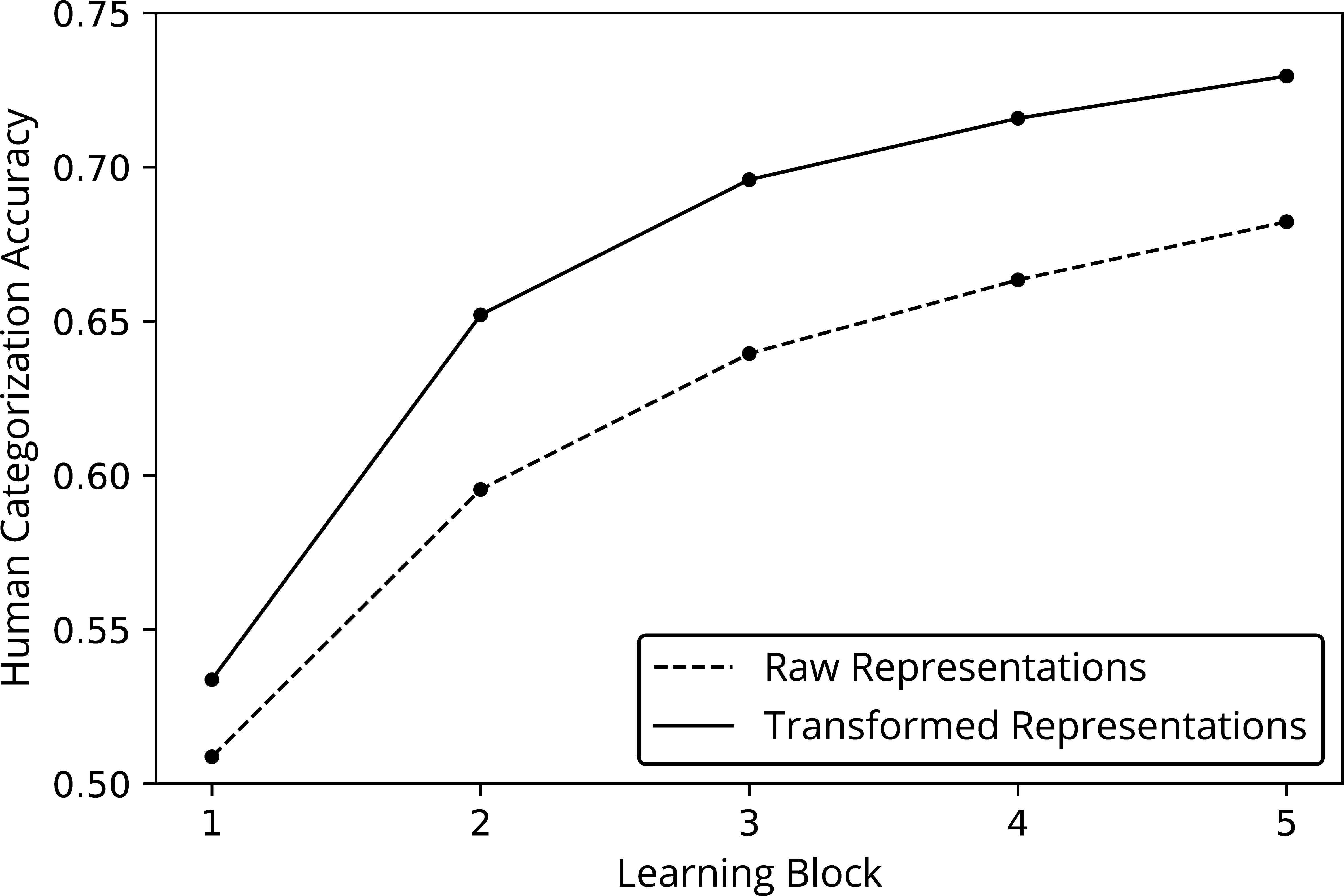}
\end{center}
\caption{Average human categorization performance for each of five learning blocks.} 
\label{fig:cat-learning-curves}
\end{figure}

Fig. \ref{fig:cat-results-extended} shows the difference in the ease with which people learned $2$-, $3$-, and $4$-category partitions derived from the raw and transformed similarities. Using DNN features, categorization performance is higher for categories derived from the transformed spaces, and a three-way ANOVA ($k \times \mbox{image set} \times \mbox{transformation}$, see Table \ref{full-anova-DNN-only}) confirmed that this effect was statistically significant ($F_{1,1404} = 66.28, p<.0001$). Participants also performed worse in the HOG+SIFT baseline condition, confirmed by a large main effect of feature set in a model including both feature sets ($F_{1,2845} = 3833.35, p<.0001$, see Table \ref{full-anova}). Notably, the effect of the transformation was reversed for the baseline features, confirmed by a significant interaction between feature set and transformation ($F_{5,2845} = 65.22, p<.0001$, see Table \ref{full-anova-baselines-only}), indicating that HOG+SIFT feature tuning may not generalize, in sharp contrast with the DNN features. To assess learning effects, we grouped trials into five learning blocks. Average learning curves for the experiments using DNN features are shown in Fig. \ref{fig:cat-learning-curves}. An ANOVA with learning block as a factor in Table \ref{anova-learning-effects} confirms a large main effect of block ($F_{4,5616} = 752.91, p<.0001$), and an interaction between block and transformation ($F_{4,5616} = 5.96, p<.0001$), likely due to the more rapid increase in performance in the first block for the transformed representation condition.

\begin{table}[]
\centering
\caption{ANOVA results for Experiment 2 using only DNN features.}
\label{full-anova-DNN-only}
\begin{tabular}{l l S[table-format=4.2] l}
                               & $df$ & $F$        & $p$         \\
\midrule
k                              & 2  & 614.95 & < 0.0001         \\
image set                      & 5  & 137.52 & < 0.0001        \\
transformation                 & 1  & 66.28 & < 0.0001         \\
k $\times$ image set                  & 10 & 7.14 & < 0.0001         \\
k $\times$ transformation             & 2  & 3.42 & < 0.01          \\
image set $\times$ transformation     & 5  & 29.20 & < 0.0001          \\
k $\times$ image set $\times$ transformation & 10 & 3.17 & < 0.001  
\end{tabular}
\end{table}

\begin{table}[]
\centering
\caption{ANOVA results for Experiment 2 using feature set as a factor.}
\label{full-anova}
\begin{tabular}{l l S[table-format=4.2] l}
                             & $df$ & $F$        & $p$         \\
\midrule
k                            & 2  & 2021.39 & < 0.0001  \\
image set                    & 5  & 169.89  & < 0.0001 \\
transformation               & 1  & 139.96 & < 0.0001  \\
feature set                  & 1  & 3833.35 & < 0.0001  \\
k $\times$ image set                & 10 & 14.96 & < 0.0001  \\
k $\times$ transformation           & 2  & 35.86 & < 0.0001  \\
k $\times$ feature set              & 2  & 13.38 & < 0.0001  \\
set $\times$ transformation         & 5  & 65.22 & < 0.0001  \\
image set $\times$ feature set      & 5  & 64.19 & < 0.0001  \\
transformation $\times$ feature set & 1  & 645.71 & < 0.0001
\end{tabular}
\end{table}

\begin{table}[]
\centering
\caption{ANOVA results for Experiment 2 using only baseline HOG+SIFT features.}
\label{full-anova-baselines-only}
\begin{tabular}{l l S[table-format=4.2] l}
                               & $df$ & $F$        & $p$         \\
\midrule
k                              & 2  & 3005.96 & < 0.0001          \\
image set                      & 5  & 108.98  & < 0.0001         \\
transformation                 & 1  & 1767.70 & < 0.0001        \\
k $\times$ image set                  & 10 & 25.67 & < 0.0001          \\
k $\times$ transformation             & 2  & 101.38 & < 0.0001          \\
image set $\times$ transformation     & 5  & 123.82 & < 0.0001        \\
k $\times$ image set $\times$ transformation & 10 & 27.85 & < 0.0001 
\end{tabular}
\end{table}

\begin{table}[]
\centering
\caption{ANOVA results for Experiment 2 using only DNN features and learning block as a factor.}
\label{anova-learning-effects}
\begin{tabular}{l l S[table-format=4.2] l}
                               & $df$ & $F$        & $p$         \\
\midrule
k                    & 2   & 605.49 & < 0.0001 \\
image set            & 5   & 137.10 & < 0.0001 \\
transformation       & 1   & 66.86  & < 0.0001 \\
block                & 4   & 752.91 & < 0.0001 \\
k $\times$ image set & 10  & 7.23   & < 0.0001 \\
k $\times$ transformation     & 2   & 3.68   & < 0.001  \\
k $\times$ block     & 8   & 39.32  & < 0.0001 \\
image set $\times$ transformation   & 5   & 29.17  & < 0.0001 \\
image set $\times$ block   & 20  & 9.51   & < 0.0001 \\
transformation $\times$ block & 4   & 5.96   & < 0.0001
\end{tabular}
\end{table}

\section{General Discussion}

The framework presented here, inspired by classic psychological methods, is the first comprehensive comparison between modern deep neural networks and human psychological representations. These artificial neural networks appear to make surprisingly good approximations to human similarities.  Importantly, they also diverge in systematic ways (e.g., lacking taxonomic representational information) \citep{mur_human_2013}. However, the representations formed by these networks can easily be transformed to produce extremely good predictions of human similarity judgments for natural images. The resulting models transfer to new stimuli, and can be used to predict complex behaviors such as the ease of category learning. Since these representations and artificial networks are easy and cheap to manipulate, they present a valuable resource for rapidly probing and mimicking human-like representations and a potential path towards studying human cognition using more naturalistic stimuli.

Were these deep representations different enough from humans (i.e., requiring nonlinear transformations and therefore additional complex feature learning), adapting them to people would require either vastly more human judgments or significantly revised network architectures, the former being quite costly and the latter presenting a massive search problem. The method we propose to transform representations is extremely effective despite being constrained to a simple reweighting of the features. The linear transformation learned can be interpreted as an analogue of dimensional attention \citep{nosofsky1987attention},  highlighting the possibility that the gap between these two sets of representations may be even smaller than we think. In fact, given that our stimulus sets are mostly restricted to single domains (e.g., fruits), whereas the DNN classifiers make all output discriminations with respect to 1000 highly diverse object classes, one would expect that certain features should become more salient, while still others should be suppressed when making judgments in context (an important real-life situation not often incorporated in machine learning models). Finally, the ability of these adapted representations to predict human categorization behavior with novel stimuli demonstrates their applicability to studying downstream cognitive processes that rely on these representations, and may have applications in the optimal design of learning software.

The proliferation of machine learning methods for representing complex stimuli is likely to continue. We see our approach as a way to leverage these advances and combine them with decades of research on psychological methods to shed light on deep questions about human cognition. This allows us to learn something about the potential weaknesses in artificial systems, and inspires new ideas for engineering those systems to more closely match human abilities. Most significantly, it provides a way for psychologists to begin to answer questions about the exercise of intelligence in a complex world, abstracting over the representational challenges that can make it difficult to identify higher-level principles of cognition in natural settings.

\setcounter{secnumdepth}{0}

\bibliography{main}

\end{document}